\documentclass{article}

\usepackage{arxiv}

\usepackage[utf8]{inputenc} 
\usepackage[T1]{fontenc}    
\usepackage{hyperref}       
\usepackage{url}            
\usepackage{booktabs}       
\usepackage{amsfonts}       
\usepackage{nicefrac}       
\usepackage{microtype}      
\usepackage{lipsum}
\usepackage{svg}
\usepackage{graphicx}
\graphicspath{ {./images/} }
\usepackage{amsmath} 
\usepackage{pdflscape}
\usepackage{xcolor}
\usepackage[normalem]{ulem}
\usepackage{comment}
 \usepackage{longtable}
\usepackage{tabularx}
\usepackage{caption} 
\usepackage{subcaption}
\usepackage{longtable}
\usepackage{booktabs}
\usepackage[utf8]{inputenc}

\title{YOLOv11: An Overview of the Key Architectural Enhancements}

\author{
  \textbf{Rahima Khanam}\textsuperscript{*} and \textbf{Muhammad Hussain}\\[1ex] 
  \begin{minipage}[t]{0.90\textwidth}
    \scriptsize Department of Computer Science, Huddersfield University, Queensgate, Huddersfield HD1 3DH, UK; \\
    \textsuperscript{*}Correspondence: rahima.khanam@hud.ac.uk;
  \end{minipage}
}
\begin{document}
\maketitle
\begin{abstract}
This study presents an architectural analysis of YOLOv11, the latest iteration in the YOLO (You Only Look Once) series of object detection models. We examine the models architectural innovations, including the introduction of the C3k2 (Cross Stage Partial with kernel size 2) block, SPPF (Spatial Pyramid Pooling - Fast), and C2PSA (Convolutional block with Parallel Spatial Attention) components, which contribute in improving the models performance in several ways such as enhanced feature extraction. The paper explores YOLOv11's expanded capabilities across various computer vision tasks, including object detection, instance segmentation, pose estimation, and oriented object detection (OBB). We review the model's performance improvements in terms of mean Average Precision (mAP) and computational efficiency compared to its predecessors, with a focus on the trade-off between parameter count and accuracy. Additionally, the study discusses YOLOv11's versatility across different model sizes, from nano to extra-large, catering to diverse application needs from edge devices to high-performance computing environments. Our research provides insights into YOLOv11's position within the broader landscape of object detection and its potential impact on real-time computer vision applications.
\end{abstract}


\section{Introduction}

Computer vision, a rapidly advancing field, enables machines to interpret and understand visual data~\cite{sonka2013image}. A crucial aspect of this domain is object detection\cite{zou2023object}, which involves the precise identification and localization of objects within images or video streams\cite{zhao2019object}. Recent years have witnessed remarkable progress in algorithmic approaches to address this challenge~\cite{hussain2024depth}.

A pivotal breakthrough in object detection came with the introduction of the You Only Look Once (YOLO) algorithm by Redmon et al. in 2015~\cite{redmon2016you}. This innovative approach, as its name suggests, processes the entire image in a single pass to detect objects and their locations. YOLO's methodology diverges from traditional two-stage detection processes by framing object detection as a regression problem~\cite{redmon2016you}. It employs a single convolutional neural network to simultaneously predict bounding boxes and class probabilities across the entire image~\cite{du2018understanding}, streamlining the detection pipeline compared to more complex traditional methods.

YOLOv11 is the latest iteration in the YOLO series, building upon the foundation established by YOLOv1. Unveiled at the YOLO Vision 2024 (YV24) conference, YOLOv11 represents a significant leap forward in real-time object detection technology. This new version introduces substantial enhancements in both architecture and training methodologies, pushing the boundaries of accuracy, speed, and efficiency.

YOLOv11's innovative design incorporates advanced feature extraction techniques, allowing for more nuanced detail capture while maintaining a lean parameter count. This results in improved accuracy across a diverse range of computer vision (CV) tasks, from object detection to classification. Furthermore, YOLOv11 achieves remarkable gains in processing speed, substantially enhancing real-time performance capabilities.

In the following sections, this paper will provide a comprehensive analysis of YOLOv11's architecture, exploring its key components and innovations. We will examine the evolution of YOLO models, leading up to the development of YOLOv11. The study will delve into the model's expanded capabilities across various CV tasks, including object detection, instance segmentation, pose estimation, and oriented object detection. We will also review YOLOv11's performance improvements in terms of accuracy and computational efficiency compared to its predecessors, with a particular focus on its versatility across different model sizes. Finally, we will discuss the potential impact of YOLOv11 on real-time CV applications and its position within the broader landscape of object detection technologies.


\section{Evolution of YOLO models}
Table \ref{tab:yolo_versions} illustrates the progression of YOLO models from their inception to the most recent versions. Each iteration has brought significant improvements in object detection capabilities, computational efficiency, and versatility in handling various CV tasks.

\begin{table}[ht]
\centering
\caption{YOLO: Evolution of models}
\label{tab:yolo_versions}
\begin{tabular}{|l|l|p{5cm}|p{5cm}|l|}
\hline
Release & Year & Tasks & Contributions & Framework \\
\hline
YOLO \cite{redmon2016you} & 2015 & Object Detection, Basic Classification & Single-stage object detector & Darknet \\
YOLOv2 \cite{redmon2017yolo9000} & 2016 & Object Detection, Improved Classification & Multi-scale training, dimension clustering & Darknet \\
YOLOv3 \cite{redmon2018yolov3} & 2018 & Object Detection, Multi-scale Detection & SPP block, Darknet-53 backbone & Darknet \\
YOLOv4 \cite{bochkovskiy2020yolov4} & 2020 & Object Detection, Basic Object Tracking & Mish activation, CSPDarknet-53 backbone & Darknet \\
YOLOv5 \cite{yolov5_blog} & 2020 & Object Detection, Basic Instance Segmentation (via custom modifications) & Anchor-free detection, SWISH activation, PANet & PyTorch \\
YOLOv6 \cite{li2022yolov6} & 2022 & Object Detection, Instance Segmentation & Self-attention, anchor-free OD & PyTorch \\
YOLOv7 \cite{wang2023yolov7} & 2022 & Object Detection, Object Tracking, Instance Segmentation & Transformers, E-ELAN reparameterisation & PyTorch \\
YOLOv8 \cite{Solawetz2023yolov8} & 2023 & Object Detection, Instance Segmentation, Panoptic Segmentation, Keypoint Estimation & GANs, anchor-free detection & PyTorch \\
YOLOv9 \cite{wang2024yolov9} & 2024 & Object Detection, Instance Segmentation & PGI and GELAN & PyTorch \\
YOLOv10 \cite{wang2024yolov10} & 2024 & Object Detection & Consistent dual assignments for NMS-free training & PyTorch \\
\hline
\end{tabular}
\end{table}

This evolution showcases the rapid advancement in object detection technologies, with each version introducing novel features and expanding the range of supported tasks. From the original YOLO's groundbreaking single-stage detection to YOLOv10's NMS-free training, the series has consistently pushed the boundaries of real-time object detection.

The latest iteration, YOLO11, builds upon this legacy with further enhancements in feature extraction, efficiency, and multi-task capabilities. Our subsequent analysis will delve into YOLO11's architectural innovations, including its improved backbone and neck structures, and its performance across various computer vision tasks such as object detection, instance segmentation, and pose estimation.

\section{What is YOLOv11?}
The evolution of the YOLO algorithm reaches new heights with the introduction of YOLOv11 \cite{yolo11_ultralytics}, representing a significant advancement in real-time object detection technology. This latest iteration builds upon the strengths of its predecessors while introducing novel capabilities that expand its utility across diverse CV applications.

YOLOv11 distinguishes itself through its enhanced adaptability, supporting an expanded range of CV tasks beyond traditional object detection. Notable among these are posture estimation and instance segmentation, broadening the model's applicability in various domains. YOLOv11's design focuses on balancing power and practicality, aiming to address specific challenges across various industries with increased accuracy and efficiency.

This latest model demonstrates the ongoing evolution of real-time object detection technology, pushing the boundaries of what's possible in CV applications. Its versatility and performance improvements position YOLOv11 as a significant advancement in the field, potentially opening new avenues for real-world implementation across diverse sectors.

\section{Architectural footprint of Yolov11}

The YOLO framework revolutionized object detection by introducing a unified neural network architecture that simultaneously handles both bounding box regression and object classification tasks \cite{khanam2024comprehensive}. This integrated approach marked a significant departure from traditional two-stage detection methods, offering end-to-end training capabilities through its fully differentiable design.

At its core, the YOLO architecture consists of three fundamental components. First, the \textbf{backbone} serves as the primary feature extractor, utilizing convolutional neural networks to transform raw image data into multi-scale feature maps. Second, the \textbf{neck} component acts as an intermediate processing stage, employing specialized layers to aggregate and enhance feature representations across different scales. Third, the \textbf{head} component functions as the prediction mechanism, generating the final outputs for object localization and classification based on the refined feature maps.

Building on this established architecture, YOLO11 extends and enhances the foundation laid by YOLOv8, introducing architectural innovations and parameter optimizations to achieve superior detection performance as illustrated in Figure \ref{fig:yolo-arch}. The following sections detail the key architectural modifications implemented in YOLO11:

\begin{figure}[h]
    \centering
    \includegraphics[width=0.7\linewidth]{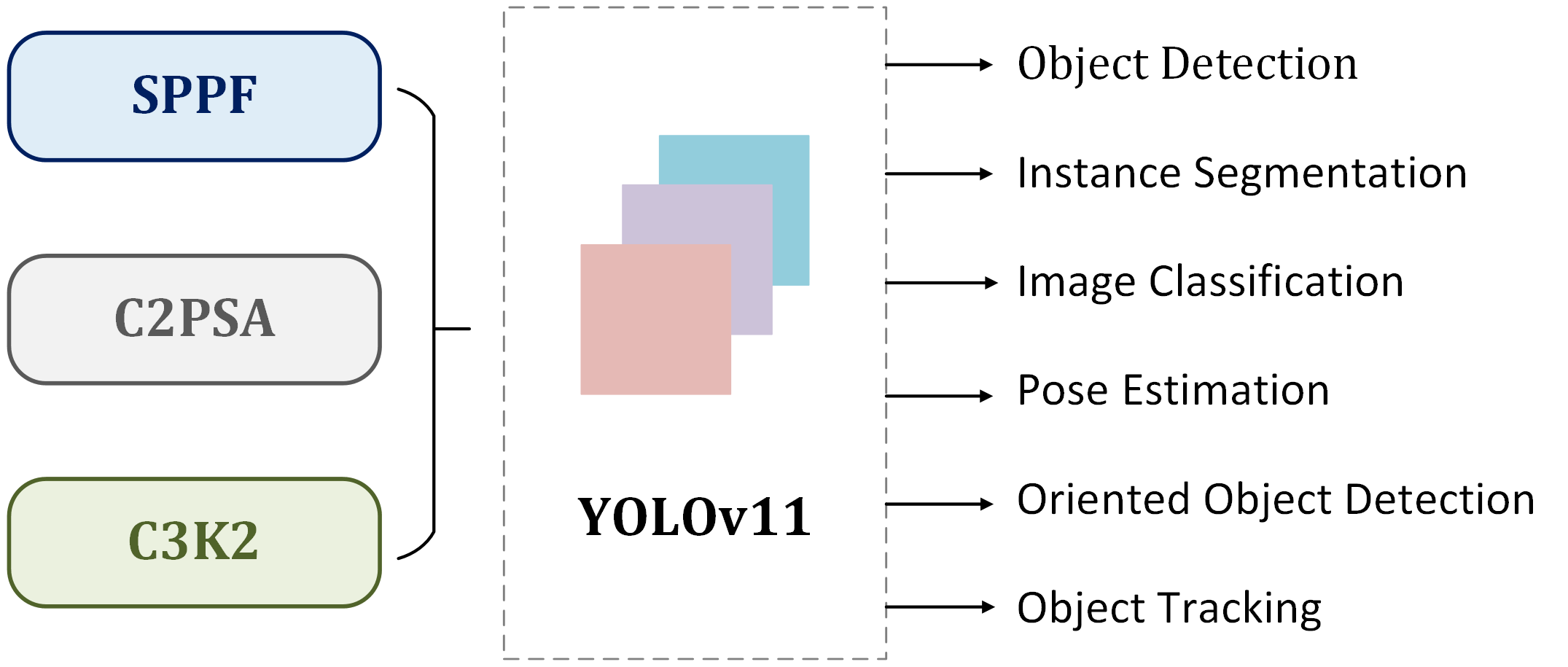}
    \caption{Key architectural modules in YOLO11}
    \label{fig:yolo-arch}
\end{figure}

\subsection{Backbone}
The backbone is a crucial component of the YOLO architecture, responsible for extracting features from the input image at multiple scales. This process involves stacking convolutional layers and specialized blocks to generate feature maps at various resolutions.

\subsubsection{Convolutional Layers}
YOLOv11 maintains a structure similar to its predecessors, utilizing initial convolutional layers to downsample the image. These layers form the foundation of the feature extraction process, gradually reducing spatial dimensions while increasing the number of channels. A significant improvement in YOLO11 is the introduction of the C3k2 block, which replaces the C2f block used in previous versions \cite{opencvyolov11}. The C3k2 block is a more computationally efficient implementation of the Cross Stage Partial (CSP) Bottleneck. It employs two smaller convolutions instead of one large convolution, as seen in YOLOv8 \cite{Solawetz2023yolov8}. The "k2" in C3k2 indicates a smaller kernel size, which contributes to faster processing while maintaining performance.

\subsubsection{SPPF and C2PSA}
YOLO11 retains the Spatial Pyramid Pooling - Fast (SPPF) block from previous versions but introduces a new Cross Stage Partial with Spatial Attention (C2PSA) block after it \cite{opencvyolov11}. The C2PSA block is a notable addition that enhances spatial attention in the feature maps. This spatial attention mechanism allows the model to focus more effectively on important regions within the image. By pooling features spatially, the C2PSA block enables YOLO11 to concentrate on specific areas of interest, potentially improving detection accuracy for objects of varying sizes and positions.

\subsection{Neck}
The neck combines features at different scales and transmits them to the head for prediction. This process typically involves upsampling and concatenation of feature maps from different levels, enabling the model to capture multi-scale information effectively.

\subsubsection{C3k2 Block}
YOLO11 introduces a significant change by replacing the C2f block in the neck with the C3k2 block. The C3k2 block is designed to be faster and more efficient, enhancing the overall performance of the feature aggregation process. After upsampling and concatenation, the neck in YOLO11 incorporates this improved block, resulting in enhanced speed and performance \cite{opencvyolov11}.

\subsubsection{Attention Mechanism}
A notable addition to YOLO11 is its increased focus on spatial attention through the C2PSA module. This attention mechanism enables the model to concentrate on key regions within the image, potentially leading to more accurate detection, especially for smaller or partially occluded objects. The inclusion of C2PSA sets YOLO11 apart from its predecessor, YOLOv8, which lacks this specific attention mechanism \cite{opencvyolov11}.

\subsection{Head}
The head of YOLOv11 is responsible for generating the final predictions in terms of object detection and classification. It processes the feature maps passed from the neck, ultimately outputting bounding boxes and class labels for objects within the image.

\subsubsection{C3k2 Block}
In the head section, YOLOv11 utilizes multiple C3k2 blocks to efficiently process and refine the feature maps. The C3k2 blocks are placed in several pathways within the head, functioning to process multi-scale features at different depths. The C3k2 block exhibits flexibility depending on the value of the c3k parameter:

\begin{itemize}
    \item When c3k = False, the C3k2 module behaves similarly to the C2f block, utilizing a standard bottleneck structure.
    \item When c3k = True, the bottleneck structure is replaced by the C3 module, which allows for deeper and more complex feature extraction.
\end{itemize}

Key characteristics of the C3k2 block:

\begin{itemize}
    \item \textbf{Faster processing: }The use of two smaller convolutions reduces the computational overhead compared to a single large convolution, leading to quicker feature extraction.
    \item \textbf{Parameter efficiency:} C3k2 is a more compact version of the CSP bottleneck, making the architecture more efficient in terms of the number of trainable parameters.
\end{itemize}

Another notable addition is the C3k block, which offers enhanced flexibility by allowing customizable kernel sizes. The adaptability of C3k is particularly useful for extracting more detailed features from images, contributing to improved detection accuracy.

\subsubsection{CBS Blocks}
The head of YOLOv11 includes several CBS (Convolution-BatchNorm-Silu) \cite{feng2024application} layers after the C3k2 blocks. These layers further refine the feature maps by:
\begin{itemize}
    \item Extracting relevant features for accurate object detection.
    \item Stabilizing and normalizing the data flow through batch normalization.
    \item Utilizing the  Sigmoid Linear Unit (SiLU) activation function for non-linearity, which improves model performance.
\end{itemize}

CBS blocks serve as foundational components in both feature extraction and the detection process, ensuring that the refined feature maps are passed to the subsequent layers for bounding box and classification predictions.

\subsubsection{Final Convolutional Layers and Detect Layer}
Each detection branch ends with a set of Conv2D layers, which reduce the features to the required number of outputs for bounding box coordinates and class predictions. The final Detect layer consolidates these predictions, which include:

\begin{itemize}
    \item Bounding box coordinates for localizing objects in the image.
    \item Objectness scores that indicate the presence of objects.
    \item Class scores for determining the class of the detected object.
\end{itemize}

\section{Key Computer Vision Tasks Supported by YOLO11}
YOLO11 supports a diverse range of CV tasks, showcasing its versatility and power in various applications. Here's an overview of the key tasks:
\begin{enumerate}
    \item \textbf{Object Detection:} YOLO11 excels in identifying and localizing objects within images or video frames, providing bounding boxes for each detected item \cite{ultralyticsisyolov112024}. This capability finds applications in surveillance systems, autonomous vehicles, and retail analytics, where precise object identification is crucial \cite{ultralyticsyolov112024}.
    \item \textbf{Instance Segmentation:} Going beyond simple detection, YOLO11 can identify and separate individual objects within an image down to the pixel level \cite{ultralyticsisyolov112024}. This fine-grained segmentation is particularly valuable in medical imaging for precise organ or tumor delineation, and in manufacturing for detailed defect detection \cite{ultralyticsyolov112024}.
    \item \textbf{Image Classification:} YOLOv11 is capable of classifying entire images into predetermined categories, making it ideal for applications like product categorization in e-commerce platforms or wildlife monitoring in ecological studies \cite{ultralyticsyolov112024}.
    \item \textbf{Pose Estimation:} The model can detect specific key points within images or video frames to track movements or poses. This capability is beneficial for fitness tracking applications, sports performance analysis, and various healthcare applications requiring motion assessment \cite{ultralyticsyolov112024}.
    \item \textbf{Oriented Object Detection (OBB):} YOLO11 introduces the ability to detect objects with an orientation angle, allowing for more precise localization of rotated objects. This feature is especially valuable in aerial imagery analysis, robotics, and warehouse automation tasks where object orientation is crucial \cite{ultralyticsyolov112024}.
    \item \textbf{Object Tracking:} It identifies and traces the path of objects in a sequence of images or video frames\cite{ultralyticsyolov112024}. This real-time tracking capability is essential for applications such as traffic monitoring, sports analysis, and security systems.
\end{enumerate}

\begin{table}[ht]
\centering
\caption{YOLOv11 Model Variants and Tasks}
\label{tab:yolo_CVtasks}
\begin{tabular}{lp{4cm}p{2.5cm}cccc}
\hline
\textbf{Model} & \textbf{Variants} & \textbf{Task} & \textbf{Inference} & \textbf{Validation} & \textbf{Training} & \textbf{Export} \\
\hline
YOLOv11 & yolo11-nano yolo11-small yolo11-medium yolo11-large yolo11-xlarge & Detection & $\checkmark$ & $\checkmark$ & $\checkmark$ & $\checkmark$ \\
YOLOv11-seg & yolo11-nano-seg yolo11-small-seg yolo11-medium-seg yolo11-large-seg yolo11-xlarge-seg & Instance Segmentation & $\checkmark$ & $\checkmark$ & $\checkmark$ & $\checkmark$ \\
YOLOv11-pose & yolo11-nano-pose yolo11-small-pose yolo11-medium-pose yolo11-large-pose yolo11-xlarge-pose & Pose/Keypoints & $\checkmark$ & $\checkmark$ & $\checkmark$ & $\checkmark$ \\
YOLOv11-obb & yolo11-nano-obb yolo11-small-obb yolo1-medium-obb yolo11-large-obb yolo11-xlarge-obb & Oriented Detection & $\checkmark$ & $\checkmark$ & $\checkmark$ & $\checkmark$ \\
YOLOv11-cls & yolo11-nano-cls yolo11-small-cls yolo11-medium-cls yolo11-large-cls yolo11-xlarge-cls & Classification & $\checkmark$ & $\checkmark$ & $\checkmark$ & $\checkmark$ \\
\hline
\end{tabular}
\end{table}

Table \ref{tab:yolo_CVtasks} outlines the YOLOv11 model variants and their corresponding tasks. Each variant is designed for specific use cases, from object detection to pose estimation. Moreover, all variants support core functionalities like inference, validation, training, and export, making YOLOv11 a versatile tool for various CV applications.

\section{Advancements and Key Features of YOLOv11}
YOLOv11 represents a significant advancement in object detection technology, building upon the foundations laid by its predecessors, YOLOv9 and YOLOv10, which were introduced earlier in 2024. This latest iteration from Ultralytics showcases enhanced architectural designs, more sophisticated feature extraction techniques, and refined training methodologies. The synergy of YOLOv11's rapid processing, high accuracy, and computational efficiency positions it as one of the most formidable models in Ultralytics' portfolio to date \cite{visoaiyolov11}. A key strength of YOLOv11 lies in its refined architecture, which facilitates the detection of subtle details even in challenging scenarios. The model's improved feature extraction capabilities allow it to identify and process a broader range of patterns and intricate elements within images. Compared to earlier versions, YOLOv11 introduces several notable enhancements:

\begin{enumerate}
    \item \textbf{Enhanced precision with reduced complexity:} The YOLOv11m variant achieves superior mean Average Precision (mAP) scores on the COCO dataset while utilizing 22\% fewer parameters than its YOLOv8m counterpart, demonstrating improved computational efficiency without compromising accuracy \cite{modelandtasksyolov11}.
    \item \textbf{Versatility in CV tasks:} YOLOv11 exhibits proficiency across a diverse array of CV applications, including pose estimation, object recognition, image classification, instance segmentation, and oriented bounding box (OBB) detection \cite{modelandtasksyolov11}.
    \item \textbf{Optimized speed and performance: }Through refined architectural designs and streamlined training pipelines, YOLOv11 achieves faster processing speeds while maintaining a balance between accuracy and computational efficiency \cite{modelandtasksyolov11}.
    \item \textbf{Streamlined parameter count:} The reduction in parameters contributes to faster model performance without significantly impacting the overall accuracy of YOLOv11 \cite{visoaiyolov11}.
    \item \textbf{Advanced feature extraction:} YOLOv11 incorporates improvements in both its backbone and neck architectures, resulting in enhanced feature extraction capabilities and, consequently, more precise object detection \cite{modelandtasksyolov11}.
    \item \textbf{Contextual adaptability: }YOLOv11 demonstrates versatility across various deployment scenarios, including cloud platforms, edge devices, and systems optimized for NVIDIA GPUs \cite{modelandtasksyolov11}.
\end{enumerate}

\begin{figure}[t]
    \centering
    \includegraphics[width=1\linewidth]{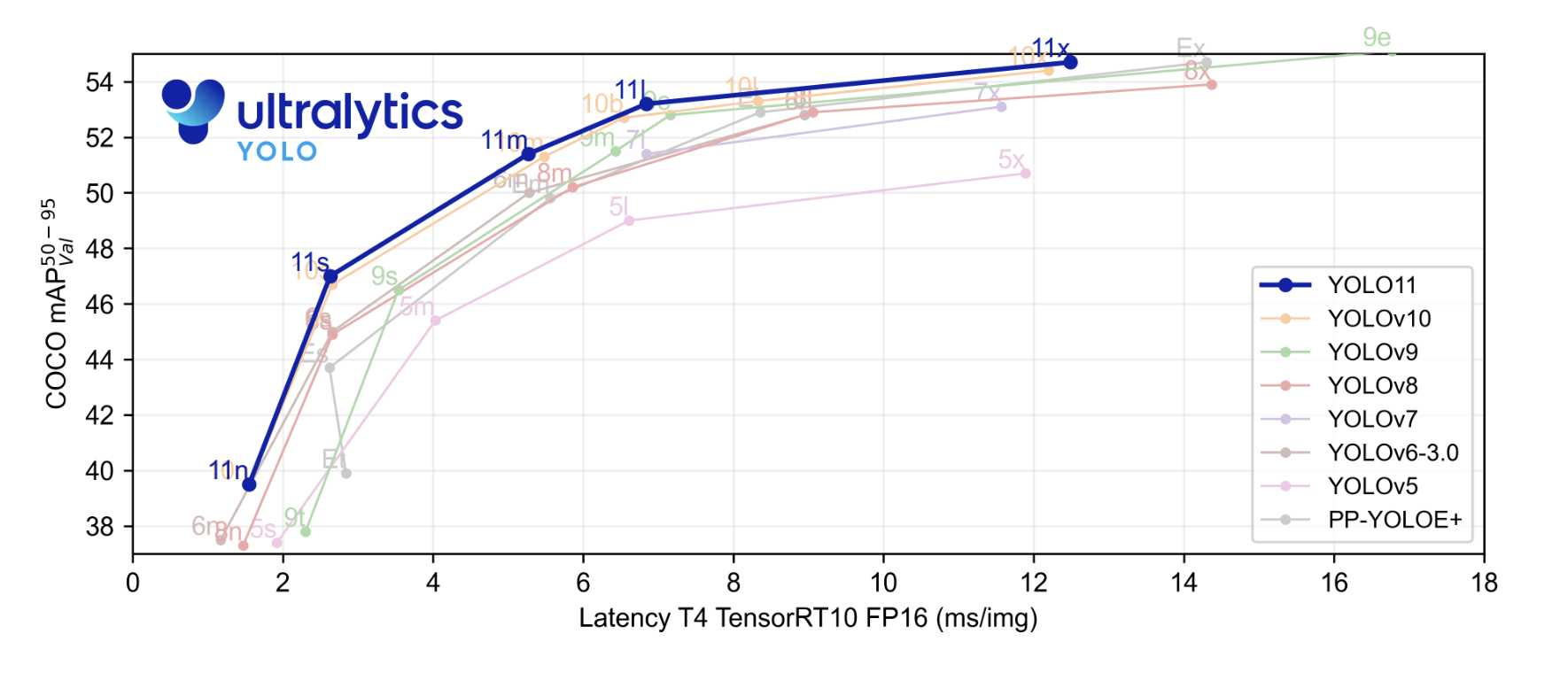}
    \caption{Benchmarking YOLOv11 Against Previous Versions \cite{modelandtasksyolov11}}
    \label{fig:yolov11-vs-prev}
\end{figure}

YOLOv11 model demonstrates significant advancements in both inference speed and accuracy compared to its predecessors. In the benchmark analysis, YOLOv11 was compared against several of its predecessors including variants such as YOLOv5 \cite{khanam2024yolov5} through to the more recent variants such as YOLOv10. As presented in Figure \ref{fig:yolov11-vs-prev}, YOLOv11 consistently outperforms these models, achieving superior mAP on the COCO dataset while maintaining a faster inference rate \cite{digitalocean_yolov11}.

The performance comparison graph depicted in Figure \ref{fig:yolov11-vs-prev} overs several key insights. The YOLOv11 variants (11n, 11s, 11m, and 11x) form a distinct performance frontier, with each model achieving higher COCO mAP$^{50-95}$ scores at their respective latency points. Notably, the YOLOv11x achieves approximately 54.5\% mAP$^{50-95}$
 at 13ms latency, surpassing all previous YOLO iterations. The intermediate variants, particularly YOLOv11m, demonstrate exceptional efficiency by achieving comparable accuracy to larger models from previous generations while requiring significantly less processing time.
 
A particularly noteworthy observation is the performance leap in the low-latency regime (2-6ms), where YOLOv11s maintains high accuracy (approximately 47\% mAP$^{50-95}$) while operating at speeds previously associated with much less accurate models. This represents a crucial advancement for real-time applications where both speed and accuracy are critical. The improvement curve of YOLOv11 also shows better scaling characteristics across its model variants, suggesting more efficient utilization of additional computational resources compared to previous generations.

\section{Discussion}
YOLO11 marks a significant leap forward in object detection technology, building upon its predecessors while introducing innovative enhancements. This latest iteration demonstrates remarkable versatility and efficiency across various CV tasks.

\begin{enumerate}
    \item \textbf{Efficiency and Scalability:} YOLO11 introduces a range of model sizes, from nano to extra-large, catering to diverse application needs. This scalability allows for deployment in scenarios ranging from resource-constrained edge devices to high-performance computing environments. The nano variant, in particular, showcases impressive speed and efficiency improvements over its predecessor, making it ideal for real-time applications.
    \item \textbf{Architectural Innovations:} The model incorporates novel architectural elements that enhance its feature extraction and processing capabilities. The incorporation of novel elements such as the C3k2 block, SPPF, and C2PSA contributes to more effective feature extraction and processing. These enhancements allow the model to better analyze and interpret complex visual information, potentially leading to improved detection accuracy across various scenarios.
    \item \textbf{Multi-Task Proficiency:} YOLO11's versatility extends beyond object detection, encompassing tasks such as instance segmentation, image classification, pose estimation, and oriented object detection. This multi-faceted approach positions YOLO11 as a comprehensive solution for diverse CV challenges.
    \item \textbf{Enhanced Attention Mechanisms:} A key advancement in YOLO11 is the integration of sophisticated spatial attention mechanisms, particularly the C2PSA component. This feature enables the model to focus more effectively on critical regions within an image, enhancing its ability to detect and analyze objects. The improved attention capability is especially beneficial for identifying complex or partially occluded objects, addressing a common challenge in object detection tasks. This refinement in spatial awareness contributes to YOLO11's overall performance improvements, particularly in challenging visual environments.
    \item \textbf{Performance Benchmarks:} Comparative analyses reveal YOLO11's superior performance, particularly in its smaller variants. The nano model, despite a slight increase in parameters, demonstrates enhanced inference speed and frames per second (FPS) compared to its predecessor. This improvement suggests that YOLO11 achieves a favorable balance between computational efficiency and detection accuracy.
    \item \textbf{Implications for Real-World Applications:} The advancements in YOLO11 have significant implications for various industries. Its improved efficiency and multi-task capabilities make it particularly suitable for applications in autonomous vehicles, surveillance systems, and industrial automation. The model's ability to perform well across different scales also opens up new possibilities for deployment in resource-constrained environments without compromising on performance.
\end{enumerate}

\section{Conclusion}
YOLOv11 represents a significant advancement in the field of CV, offering a compelling combination of enhanced performance and versatility. This latest iteration of the YOLO architecture demonstrates marked improvements in accuracy and processing speed, while simultaneously reducing the number of parameters required. Such optimizations make YOLOv11 particularly well-suited for a wide range of applications, from edge computing to cloud-based analysis.

The model's adaptability across various tasks, including object detection, instance segmentation, and pose estimation, positions it as a valuable tool for diverse industries such as emotion detection \cite{hussain2023child}, healthcare \cite{aydin2023domain} and various other industries \cite{khanam2024comprehensive}. Its seamless integration capabilities and improved efficiency make it an attractive option for businesses seeking to implement or upgrade their CV systems. In summary, YOLOv11's blend of enhanced feature extraction, optimized performance, and broad task support establishes it as a formidable solution for addressing complex visual recognition challenges in both research and practical applications.

\vspace{6pt} 






\bibliographystyle{unsrt}  
\bibliography{ref}  

\end{document}